%% file: multiagentdeep.tex
\documentclass{article} 
\usepackage{nips13submit_e,times}
\usepackage{hyperref}
\usepackage{url}
\usepackage{amsfonts}
\usepackage{amsmath}
\usepackage{algorithm}
\usepackage{algorithmicx}
\usepackage{algpseudocode}
\usepackage{graphicx}
\usepackage[colorinlistoftodos]{todonotes}
\usepackage{tabularx}
\usepackage{caption}
\usepackage{subcaption}
\usepackage{cleveref}
\usepackage{marvosym}
\usepackage{wrapfig}
\usepackage{titlesec}
\usepackage{array}
\usepackage[margin=0.5cm]{caption}
\usepackage[font=small]{caption}

\titlespacing{\paragraph}{%
  0pt}{
  0.5\baselineskip}{
  0.33em}

\title{Multiagent Cooperation and Competition with Deep Reinforcement Learning}

\author{
Ardi Tampuu*\hspace{0.3cm}
Tambet Matiisen* \hspace{0.3cm}\\
\\
{\bf
Dorian Kodelja \hspace{0.3cm}
Ilya Kuzovkin \hspace{0.3cm} 
Kristjan Korjus}\\
\\
{\bf Juhan Aru$^\dag$ \hspace{0.3cm}
Jaan Aru \hspace{0.3cm}
Raul Vicente\textsuperscript{\Letter}}\\
\\
Computational Neuroscience Lab, Institute of Computer Science, University of Tartu\\
$^\dag$ Department of Mathematics, ETH Zurich
\\
\texttt{ardi.tampuu@ut.ee, tambet.matiisen@ut.ee, raul.vicente.zafra@ut.ee}\\
\\
\emph{* these authors contributed equally to this work}
}

\nipsfinalcopy 

\begin{document}
\maketitle


%
%
\begin{abstract}
Multiagent systems appear in most social, economical, and political situations. In the present work we extend the Deep Q-Learning Network architecture proposed by Google DeepMind to multiagent environments and investigate how two agents controlled by independent Deep Q-Networks interact in the classic videogame \emph{Pong}. By manipulating the classical rewarding scheme of Pong we demonstrate how competitive and collaborative behaviors emerge. Competitive agents learn to play and score efficiently. Agents trained under collaborative rewarding schemes find an optimal strategy to keep the ball in the game as long as possible. We also describe the progression from competitive to collaborative behavior. The present work demonstrates that Deep Q-Networks can become a practical tool for studying the decentralized learning of multiagent systems living in highly complex environments.
\end{abstract}

\newcommand{\ssq}{s^{(t)}}
\newcommand{\asq}{a^{(t)}}
\newcommand{\pasq}{a^{(t-1)}}
\newcommand{\stpn}{s_{t+1}}
\newcommand{\qpi}{Q^{\pi}\left(\ssq,a_t\right)}
\newcommand{\vpi}{V^{\pi}\left(\ssq\right)}
\newcommand{\aqpi}{\hat{Q}^{\pi}}

\newcommand{\prob}[1]{\mathbb{P} \left[ #1 \right]}
\newcommand{\given}{|}
\newcommand{\expect}[1]{\mathbb{E} \left[ #1 \right]}
\newcommand{\expectx}[2]{\mathbb{E}_{#1} \left[ #2 \right]}
\newcommand{\eqdef}{\stackrel{def}{=}}

%
%
\section*{Introduction}
\label{sec:introduction}
\input{intro}

%
%
\section{Methods}
\label{sec:method}
\input{method}

%
%
\section{Results}
\label{sec:results}
\input{results}

%
%
\section{Discussion}
\label{sec:discussion}
\input{discussion}

%
%
\section{Supplementary Materials}
\label{sec:appendices}
\input{appendices}

%
%
\section*{Author Contributions}

A.T., T.M., K.K. and I.K. explored the background technical framework. D.K., A.T. and T.M. developed and tested the technical framework, and analyzed the data. A.T., T.M., D.K., I.K., Ja.A. and R.V. wrote the manuscript. All authors designed research, discussed the experiments, results, and implications, and commented on the manuscript. R.V. and Ja.A. managed the project. R.V. conceptualized the project.

\section*{Acknowledgements}
We would like to thank Andres Laan and Taivo Pungas for fruitful early discussions about the project. We gratefully acknowledge the support of NVIDIA Corporation with the donation of one GeForce GTX TITAN X GPU used for this research. We also thank the financial support of the Estonian Research Council via the grant PUT438.

%
%
\bibliographystyle{alpha}
\bibliography{refs.bib}

\end{document}

%% file: intro.tex





In the ever-changing world biological and engineered agents need to cope with unpredictability. By learning from trial-and-error an animal or a robot can adapt its behavior in a novel or changing environment. This is the main intuition behind reinforcement learning \cite{sutton1998reinforcement,poole2010artificial}. A reinforcement learning agent modifies its behavior based on the rewards that it collects while interacting with the environment. By trying to maximize the reward during these interactions an agent can learn to implement complex long-term strategies. 

Due to the astronomic number of states of any realistic scenario, for a long time algorithms implementing reinforcement learning were either limited to simple environments or needed to be assisted by additional information about the dynamics of the environment. Recently, however, the Swiss AI Lab IDSIA \cite{koutnik2013evolving} and Google DeepMind \cite{mnih2013playing,mnih2015human} have produced spectacular results in applying reinforcement learning to very high-dimensional and complex environments such as video games. In particular, DeepMind demonstrated that AI agents can achieve superhuman performance in a diverse range of Atari video games. Remarkably, the learning agent only used raw sensory input (screen images) and the reward signal (increase in game score). The proposed methodology, so called Deep Q-Network, combines a convolutional neural network for feature representation with Q-learning training \cite{lin1993reinforcement}. It represents the state-of-the-art approach for model-free reinforcement learning problems in complex environments. The fact that the same algorithm was used for learning very different games suggests its potential for general purpose applications.  

The present article builds on the work of DeepMind and explores how multiple agents controlled by autonomous Deep Q-Networks interact when sharing a complex environment. Multiagent systems appear naturally in most of social, economical and political scenarios. Indeed, most of game theory problems deal with multiple agents taking decisions to maximize their individual returns in a static environment \cite{busoniu2008comprehensive}. Collective animal behavior \cite{sumpter2010collective} and distributed control systems are also important examples of multiple autonomous systems. Phenomena such as cooperation, communication, and competition may emerge in reinforced multiagent systems.

The goal of the present work is to study emergent cooperative and competitive strategies between multiple agents controlled by autonomous Deep Q-Networks. As in the original article by DeepMind, we use Atari video games as the environment where the agents receive only the raw screen images and respective reward signals as input. We explore how two agents behave and interact in such complex environments when trained with different rewarding schemes.  

In particular, using the video game \emph{Pong} we demonstrate that by changing the rewarding schemes of the agents either competitive or cooperative behavior emerges. Competitive agents get better at scoring, while the collaborative agents find an optimal strategy to keep the ball in the game for as long as possible. We also tune the rewarding schemes in order to study the intermediate states between competitive and cooperative modes and observe the progression from competitive to collaborative behavior.


%
%

%% file: method.tex

\subsection{The Deep Q-Learning Algorithm}



The goal of reinforcement learning is to find a policy -- a rule to decide which action to take in each of the possible states -- which maximizes the agent's accumulated long term reward in a dynamical environment. The problem is especially challenging when the agent must learn without the explicit information about the dynamics of the environment or the rewards. In this case perhaps the most popular learning algorithm is Q-learning \cite{watkins1989learning}. Q-learning allows one to estimate the value or quality of an action in a particular state of the environment. 

Recently Google DeepMind trained convolutional neural networks to approximate these so-called Q-value functions. Leveraging the powerful feature representation of convolutional neural networks the so-called Deep Q-Networks have obtained state of the art results in complex environments. In particular, a trained agent achieved superhuman performance in a range of Atari video games by using only raw sensory input (screen images) and the reward signal \cite{mnih2015human,schaul2015prioritized}. 


When two or more agents share an environment the problem of reinforcement learning is much less understood. The distributed nature of the learning offers new benefits but also challenges such as the definition of good learning goals or the convergence and consistency of algorithms \cite{schwartz2014multi,busoniu2008comprehensive}. For example, in the multiagent case the environment state transitions and rewards are affected by the joint action of all agents. This means that the value of an agent's action depends on the actions of the others, and hence each agent must keep track of each of the other learning agents, possibly resulting in an ever-moving 
target. In general, learning in the presence of other agents requires a delicate trade-off between the stability and adaptive behavior of each agent. 


There exist several possible adaptations of the Q-learning algorithm for the multiagent case. However, this is an open research area and theoretical guarantees for multiagent model-free reinforcement learning algorithms are scarce and restricted to specific types of tasks \cite{schwartz2014multi,busoniu2008comprehensive}. In practice the simplest method consists of using an autonomous Q-learning algorithm for each agent in the environment, thereby using the environment as the sole source of interaction between agents. In this work we use this method due to its simplicity, decentralized nature, computational speed, and ability to produce consistent results for the range of tasks we report. Therefore, in our tests each agent is controlled by an independent Deep Q-Network with architecture and parameters as reported in \cite{mnih2015human}.

\subsection{Adaptation of the Code for the Multiplayer Paradigm}

The original code published with \cite{mnih2015human} does not provide the possibility to play multiplayer games. Whereas the agents are independent realizations of Deep Q-Networks and many of the Atari games allow multiple players, the communication protocol between the agents and the emulator restricts one to use a single player. To solve this issue we had to modify the code to allow transmitting actions from two agents, as well as receiving two sets of rewards from the environment. The game screen is fully observable and is shared between the two agents, hence no further modifications were needed to provide the state of the game to multiple agents simultaneously.

\subsection{Game Selection}


Atari Learning Environment (ALE) \cite{bellemare2012arcade} currently supports 61 games, but only a handful on them have a two player mode. In order to choose a suitable game for our multiplayer experiments we used following criteria:

\begin{enumerate}
    \item The game must have real-time two-player mode. Many games (e.g. \emph{Breakout}) alternate between two players and are therefore less suitable for our multiagent learning experiments.
    \item Deep Q-learning algorithm must be able to play the game above human level in single player mode. For example \emph{Wizard of Wor} has a two-player mode, but requires extensive labyrinth navigation, which current deep Q-learning algorithm is not able to master.
    \item The game must have a natural competitive mode. In addition, we were interested in games, where we can switch between cooperation and competition by a simple change of reward function.
\end{enumerate}


After having considered several other options we finally chose to work within the \emph{Pong} game environment because it satisfies all the criteria, it was supported by existing code and can be learned relatively quickly. It also has the advantage of being easily understood by the reader due to its simplicity and its role in the video game history.

\begin{figure}[H]
	\centering
	\includegraphics[width=0.5\textwidth]{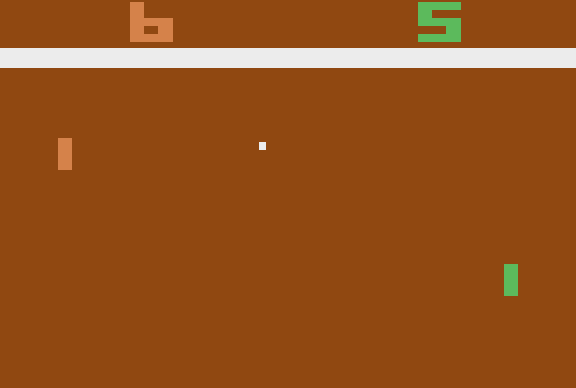}
	\caption{The \emph{Pong} game. Each agent corresponds to one of the paddles.}
    \label{fig:pong}
\end{figure}

In \emph{Pong} each agent corresponds to one of the paddles situated on the left and right side of the screen (see Figure \ref{fig:pong}). There are 4 actions that each of the two agents can take: move up, move down, stand still, and fire (to launch the ball or to start the game). Which action is taken is decided by the corresponding Deep Q-Network for both agents separately. 

While this is outside the scope of the present work, we would like to report on two other games which might mix well with interesting scientific questions on competition and collaboration. Firstly, \emph{Outlaw} is a simple shooting game that in a restricted game mode can be seen as a real-time approximation of prisoner's dilemma. Secondly, \emph{Warlords} is a game with up to four players where the emergence of collaboration in the face of an adversary could be tested. 

\subsection{Rewarding Schemes}

A central aim of this work is to study the emergence of a diversity of coordinated behaviours depending on how the agents are rewarded. Here we describe the different rewarding schemes we set for the agents. We adjust rewarding schemes by simply changing the reward both players get when the ball goes out of game. In such a way we can create several very different games within the same \emph{Pong} environment.

\subsubsection{Score More than the Opponent (Fully Competitive)}
\label{sec:competitive}

In the traditional rewarding scheme of Pong, also used in \cite{mnih2015human}, the player who last touches the outgoing ball gets a plus point, and the player losing the ball a minus point. This makes it essentially a zero-sum game, where a positive reward for the left player means negative reward for right player and vice versa. We call this fully competitive mode as the aim is to simply try to put the ball behind your opponent.

\begin{table}[H]
\renewcommand{\arraystretch}{1.5}
\centering
\begin{tabular}{ l | c c }
                        & Left player scores & Right player scores \\\hline
    Left player reward  & $+1$               & $-1$                \\
    Right player reward & $-1$               & $+1$
\end{tabular}
\caption{Rewarding scheme for the classical case of \emph{Pong}. This rewarding scheme leads to a competitive strategy.}
\label{tab:reward_competitive}
\end{table}

\subsubsection{Loosing the Ball Penalizes Both Players (Fully Cooperative)}
\label{sec:cooperative}

In this setting we want the agents to learn to keep the ball in the game for as long as possible. To achieve this, we penalize both of the players whenever the ball goes out of play. Which of the players let the ball pass does not matter and no positive rewards are given.

\begin{table}[H]
\renewcommand{\arraystretch}{1.5}
\centering
\begin{tabular}{ l | c c }
                        & Left player scores & Right player scores \\\hline
    Left player reward  & $-1$               & $-1$                \\
    Right player reward & $-1$               & $-1$
\end{tabular}
\caption{Rewarding scheme to induce cooperative strategy.}
\label{tab:reward_cooperative}
\end{table}

Notice that another possible collaborative mode would be to reward both players on each outgoing ball, but we expected this mode not to show any interesting behaviour.  

\subsubsection{Transition Between Cooperation and Competition}
\label{sec:transition}

The fully competitive and fully cooperative cases both penalize loosing the ball equally. What differentiates the two strategies are the values on the main diagonal of the reward matrix, i.e. the rewards agents get for putting the ball past the other player. If one allows this reward value to change gradually from $-1$ to $+1$, one can study intermediate scenarios that lie between competition and cooperation. We ran a series of experiments to explore the changes in the behaviour when the reward for ``scoring" a point, $\rho$, changes from $-1$ to $+1$ while the penalty for losing the ball is kept fixed at $-1$.

\begin{table}[H]
\renewcommand{\arraystretch}{1.5}
\centering
\begin{tabular}{ l | c c }
                        & Left player scores & Right player scores \\\hline
    Left player reward  & $\rho$             & $-1$                \\
    Right player reward & $-1$               & $\rho$
\end{tabular}
\caption{Rewarding scheme to explore the transition from competitive to the cooperative strategy where $\rho \in [-1,1]$.}
\label{tab:reward_transition}
\end{table}

\subsection{Training Procedure}

In all of the experiments we let the agents learn for 50 epochs \footnote{We limit the learning to 50 epochs because the Q-values predicted by the network have stabilized (see Supplementary Materials \Cref{fig:meanq_reward_all})}, 250000 time steps each. Due to using a frame skipping technique 
the agents see and select actions only on every 4th frame. In the following we use ``visible frame", ``frame" and ``time step" interchangeably.

During the training time 
, as in \cite{mnih2015human},
the exploration rate (percentage of actions chosen randomly) decreases from an initial $1.0$ to $0.05$ in million time steps and stays fixed at that value. Here, selecting a random action simply means that we draw a command randomly from amongst all the possible actions in the game instead of using the prediction that would have been made by the Deep Q-Network. A more detailed description of the training procedure and parameters can be found in \cite{mnih2015human}.

The convergence of Q-values is an indicator of the convergence of the deep Q-network controlling the behaviour of the agent.
Hence, we monitor the \emph{average maximal Q-values} of 500 randomly selected game situations, set aside before training begins. We feed these states to the networks after each training epoch and record the maximal value in the last layer of each of the Deep Q-Networks. These maximal values correspond to how highly the agent rates its best action in each of the given states and thus estimates the quality of the state itself. 

After each epoch snapshots of the Deep Q-Networks are stored to facilitate the future study of the training process.

\subsection{Collecting the Game Statistics}

To obtain quantitative measures of the agents' behaviour in the \emph{Pong} environment, we identified and counted specific events in the game, e.g. bouncing of the ball against the paddle or the wall. We used Stella \cite{mott2003stella} integrated debugger to detect these events. Stella makes it possible to identify the exact memory locations that hold the numbers of bounces made by the players as well as several other game events.

To quantitatively evaluate the emergence of cooperative or competitive behaviour we collected the measures after each training epoch for all of the rewarding schemes. After the end of each epoch we took the deep Q-networks of both players in their current state and had them play 10 games \footnote{In \emph{Pong} one game consists of multiple exchanges and lasts until one of the agents reaches 21 points} using different random seeds to obtain the statistics on the measures. During the testing phase the exploration rate was set to 0.01. The behavioral measures we used are the following:

\begin{itemize}
    \item \textbf{Average \emph{paddle-bounces} per point} counts how many times the ball bounces between two players before one of them scores a point. Randomly playing agents almost never hit the ball. Well trained agents hit the ball multiple times in an exchange. Hereafter we refer to this statistic as \emph{paddle-bounces}.
    \item \textbf{Average \emph{wall-bounces} per paddle-bounce} quantifies how many times the ball bounces from top and bottom walls before it reaches the other player. It is possible to hit the ball in an acute angle so that it bounces the walls several times before reaching the other player. Depending on the strategy, players might prefer to send the ball directly to the other player or use the wall bounces. Hereafter we refer to this statistic as \emph{wall-bounces}.
    \item \textbf{Average \emph{serving time} per point} shows how long it takes for the players to restart the game after the ball was lost (measured in frames). To restart the game, the agent who just scored has to send a specific command (fire). Depending on the rewarding scheme the agents might want to avoid restarting the game. Hereafter we refer to this statistic as \emph{serving time}.
\end{itemize}
\pagebreak

%% file: results.tex

In this section we explain our main result: the emergence of competitive and collaborative behaviors in the playing strategies of the learning agents.  

\subsection{Emergence of Competitive Agents}
In the full competitive (zero-sum) rewarding scheme each agent obtains an immediate reward when the ball gets past the other agent and an immediate punishment when it misses the ball. 

\begin{figure}[h!] 
    \centering
    \begin{subfigure}[b]{0.32\textwidth}
        \captionsetup{justification=centering}
        \includegraphics[width=\textwidth]{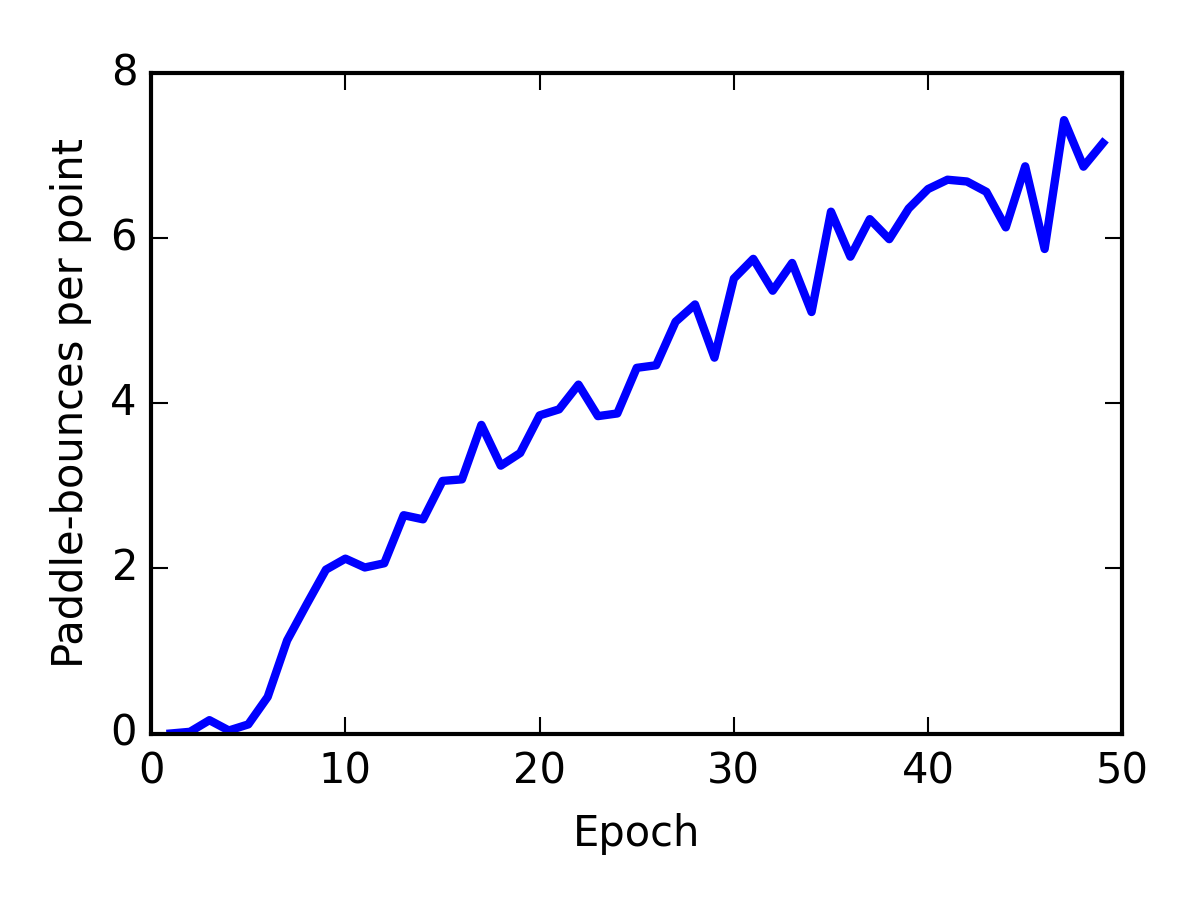}
        \caption{Paddle-bounces per point \phantom{-------------------}}
        \label{fig:sidebounces_comp}
    \end{subfigure}
    ~ 
    \begin{subfigure}[b]{0.32\textwidth}
        \captionsetup{justification=centering}
        \includegraphics[width=\textwidth]{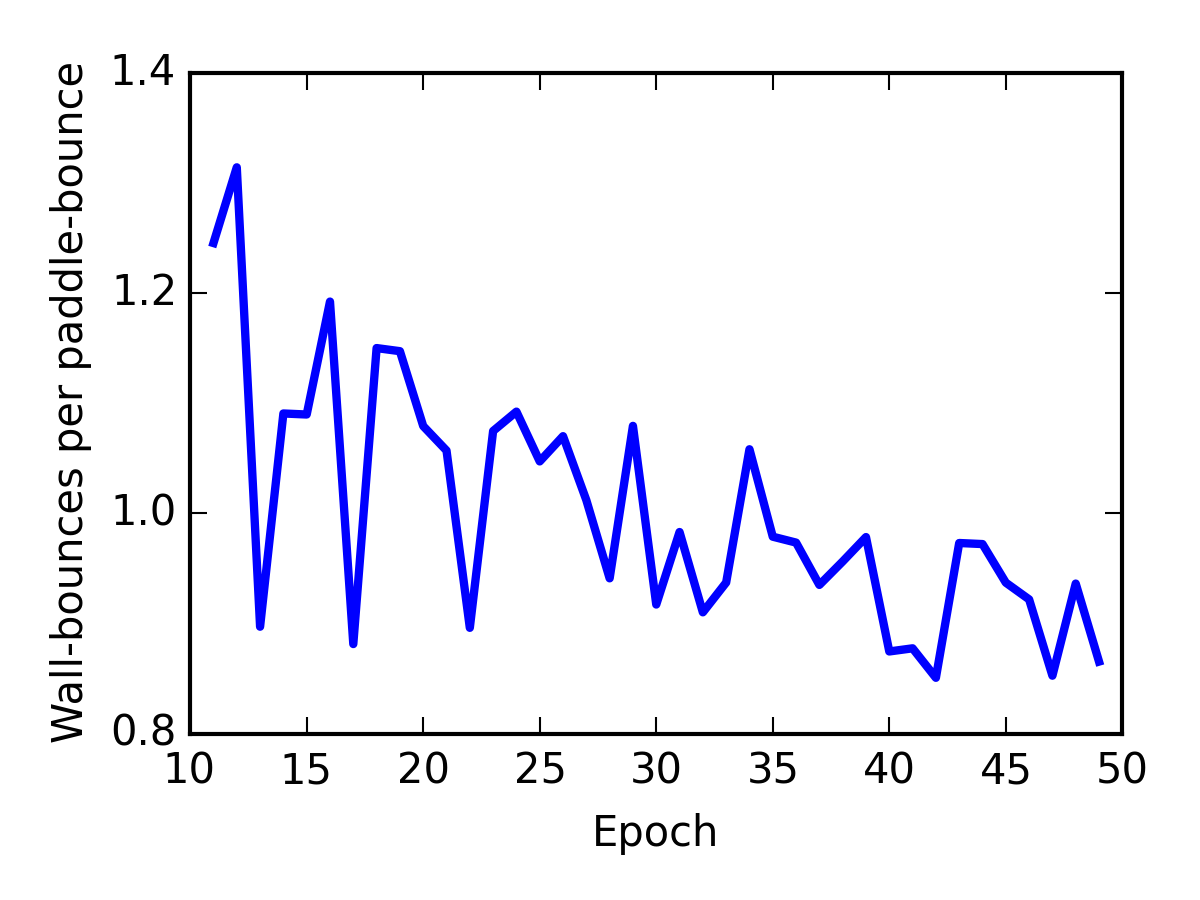}
        \caption{Wall-bounces per paddle-bounce}
        \label{fig:wallbounce_comp}
    \end{subfigure}
    ~ 
    \begin{subfigure}[b]{0.32\textwidth}
        \captionsetup{justification=centering}
        \includegraphics[width=\textwidth]{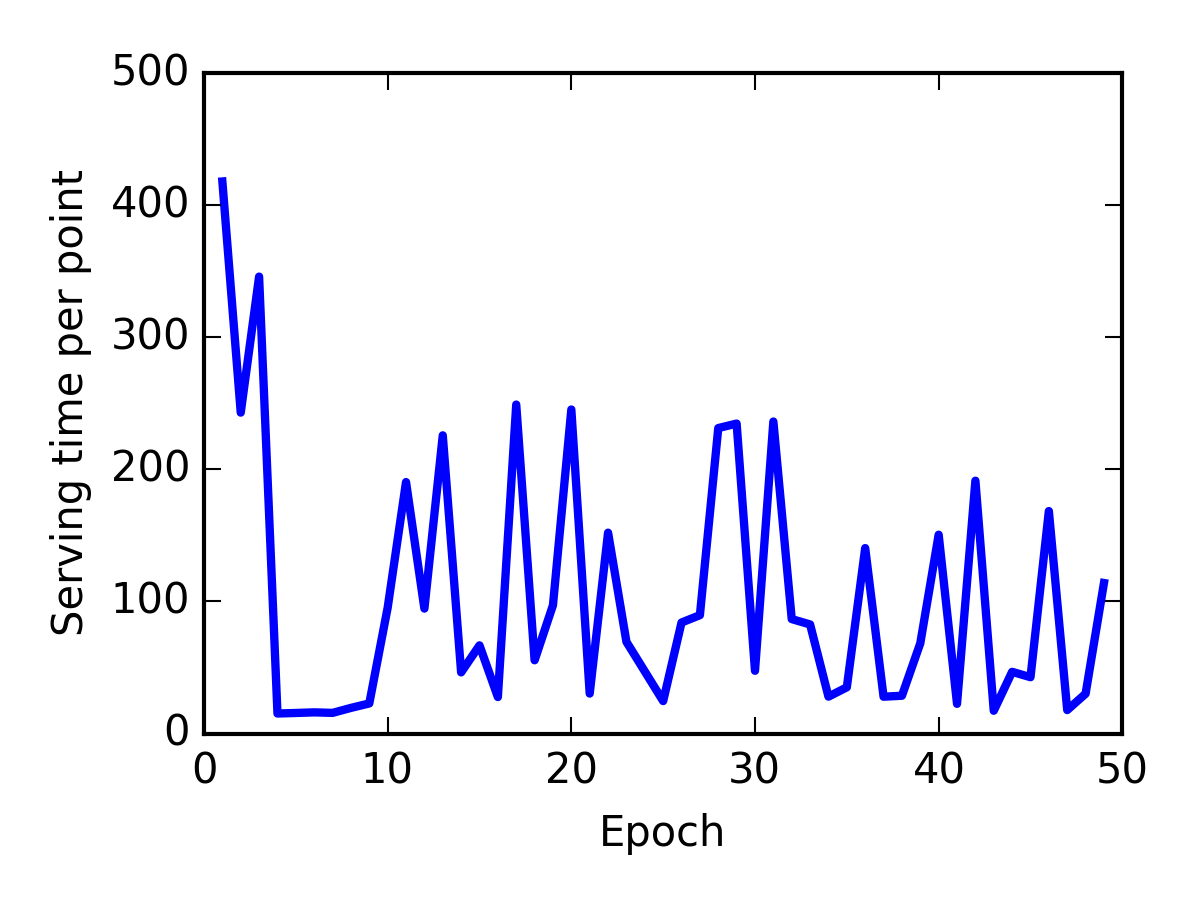}
        \caption{Serving time per point \phantom{-------------------}}
        \label{fig:serving_time_comp}

    \end{subfigure}
    \caption{Evolution of the behaviour of the competitive agents during training. (\subref{fig:sidebounces_comp}) The number of paddle-bounces increases indicating that the players get better at catching the ball. (\subref{fig:wallbounce_comp}) The frequency of the ball hitting the upper and lower walls decreases slowly with training. The first 10 epochs are omitted from the plot as very few paddle-bounces were made by the agents and the metric was very noisy. (\subref{fig:serving_time_comp}) Serving time decreases abruptly in early stages of training- the agents learn to put the ball back into play. Serving time is measured in frames. }
    \label{fig:evolution_VS}
\end{figure}

Initially the agents fail to hit the ball at all but with training both agents become more and more proficient. The learning of both agents progresses continuously. \Cref{fig:evolution_VS} summarizes the evolution of the quantitative descriptors of behaviour during training.

Qualitatively we can report that by the end of training both agents are capable of playing the game reasonably well. First of all, both players are capable of winning regardless of who was the one to serve. Secondly, the exchanges can last for a considerable amount of touches (\Cref{fig:sidebounces_comp}) even after the speed of the ball has increased. Thirdly we observe that the agents have learned to put the ball back in play rapidly (\Cref{fig:serving_time_comp}). 

\begin{figure}[h!]
\centering
\includegraphics[width=\textwidth]{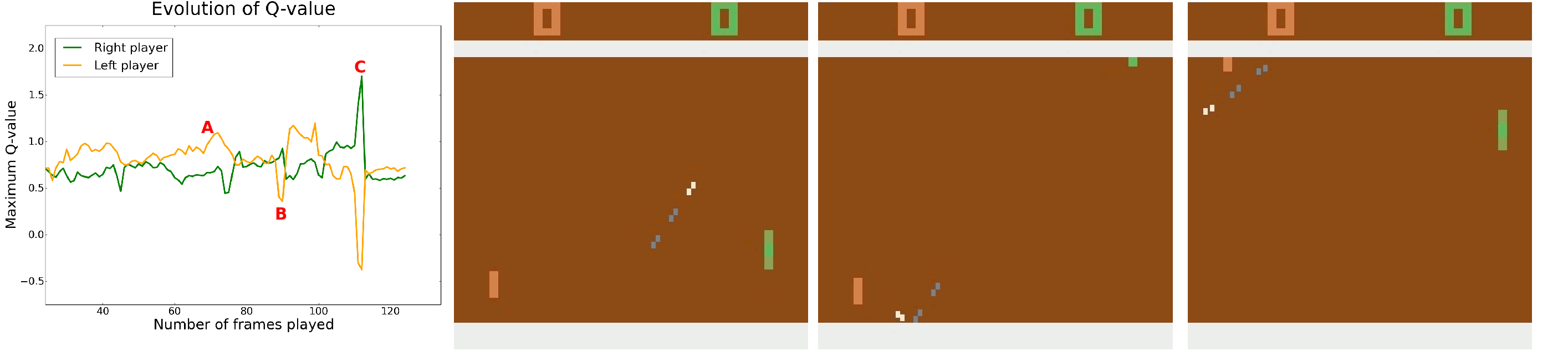}
\caption{A competitive game - game situations and the Q-values predicted by the agents. A) The left player predicts that the right player will not reach the ball as it is rapidly moving upwards. B) A change in the direction of the ball causes the left player's reward expectation to drop. C) Players understand that the ball will inevitably go out of the play. See section \ref{sec:videos} for videos illustrating other game situations and the corresponding agents' Q-values.}
\label{fig:competitive_game}
\end{figure}

\Cref{fig:competitive_game} illustrates how the agents' predictions of their rewards evolve during an exchange. A first observation is that the Q-values predicted by the agents are optimistic, both players predicting positive future rewards in most situations. The figure also demonstrates that the agents' reward expectations correlate well with game situations. More precisely, one can observe that whenever the ball is travelling towards a player its reward expectation drops and the opponent's expectation increases. This drop occurs because even a well trained agent might miss the ball (at least $5\%$ of the actions are taken randomly during training) resulting in $-1$ and $+1$ rewards for the two players respectively.

We also note that the Q-values correlate with the speed of the ball (data not shown). The faster the ball travels, the bigger are the changes in Q-values - possibly because there is less time to correct for a bad initial position or a 
random move.

Interestingly, one can also notice that as soon as the serving player has relaunched the ball its reward expectation increases slightly. This immediate increase in Q-value makes the agents choose to serve the ball as soon as possible and thus explains the decrease in serving time (\Cref{fig:serving_time_comp}).

See section \ref{sec:videos} for a short example video displaying the behavior of the agents in different game situations and the corresponding Q-values.

\subsection{Emergence of Collaborative Agents}
In the fully cooperative rewarding scheme each agent receives an immediate punishment whenever the ball get past either of the agents. Thus, the agents' are motivated to keep the ball alive. The agents get no positive rewards and the best they can achieve is to minimize the number of times the ball is lost. 
\begin{figure}[h] 
    \centering
    \begin{subfigure}[b]{0.32\textwidth}
        \captionsetup{justification=centering}
        \includegraphics[width=\textwidth]{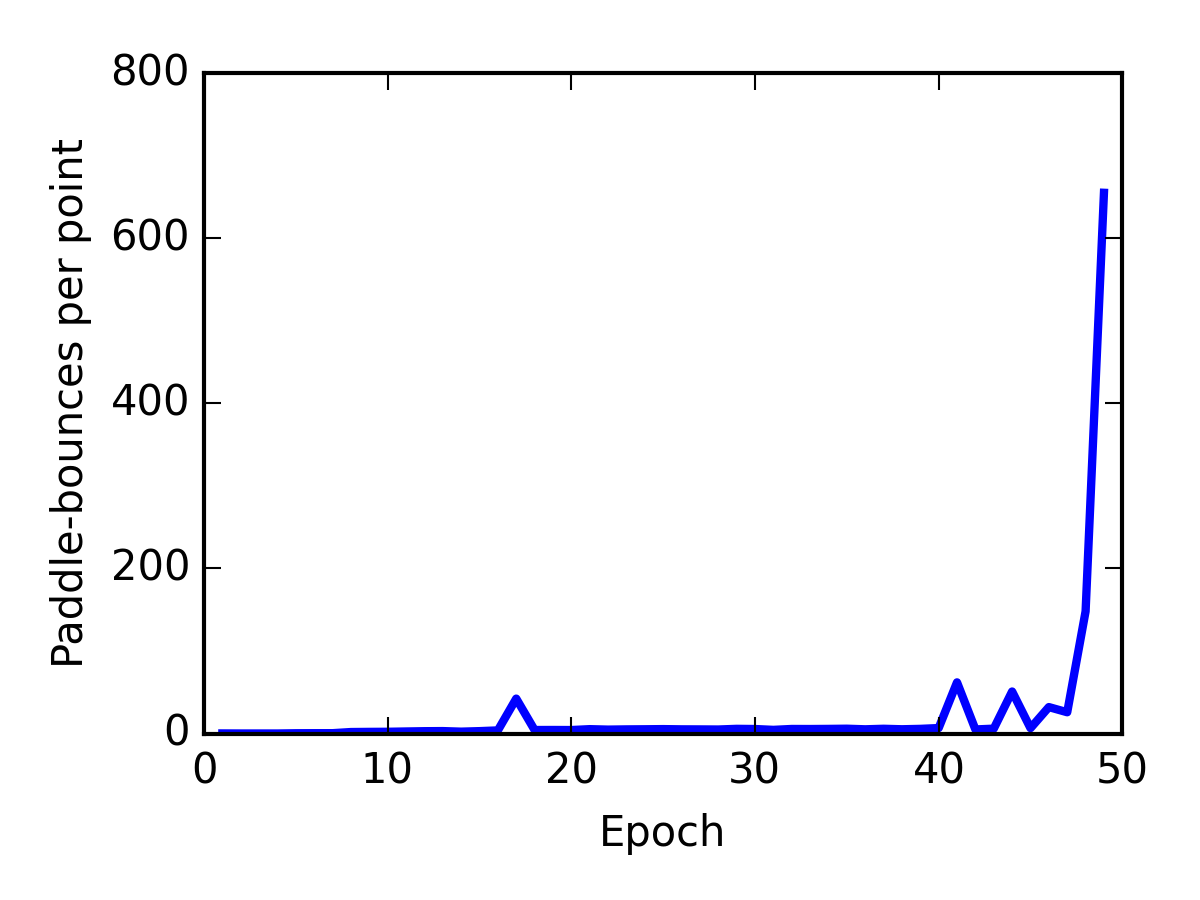}
        \caption{Paddle-bounces per point \phantom{-------------------}}
        \label{fig:sidebounces_coop}
    \end{subfigure}
    ~ 
    \begin{subfigure}[b]{0.32\textwidth}
        \captionsetup{justification=centering}
        \includegraphics[width=\textwidth]{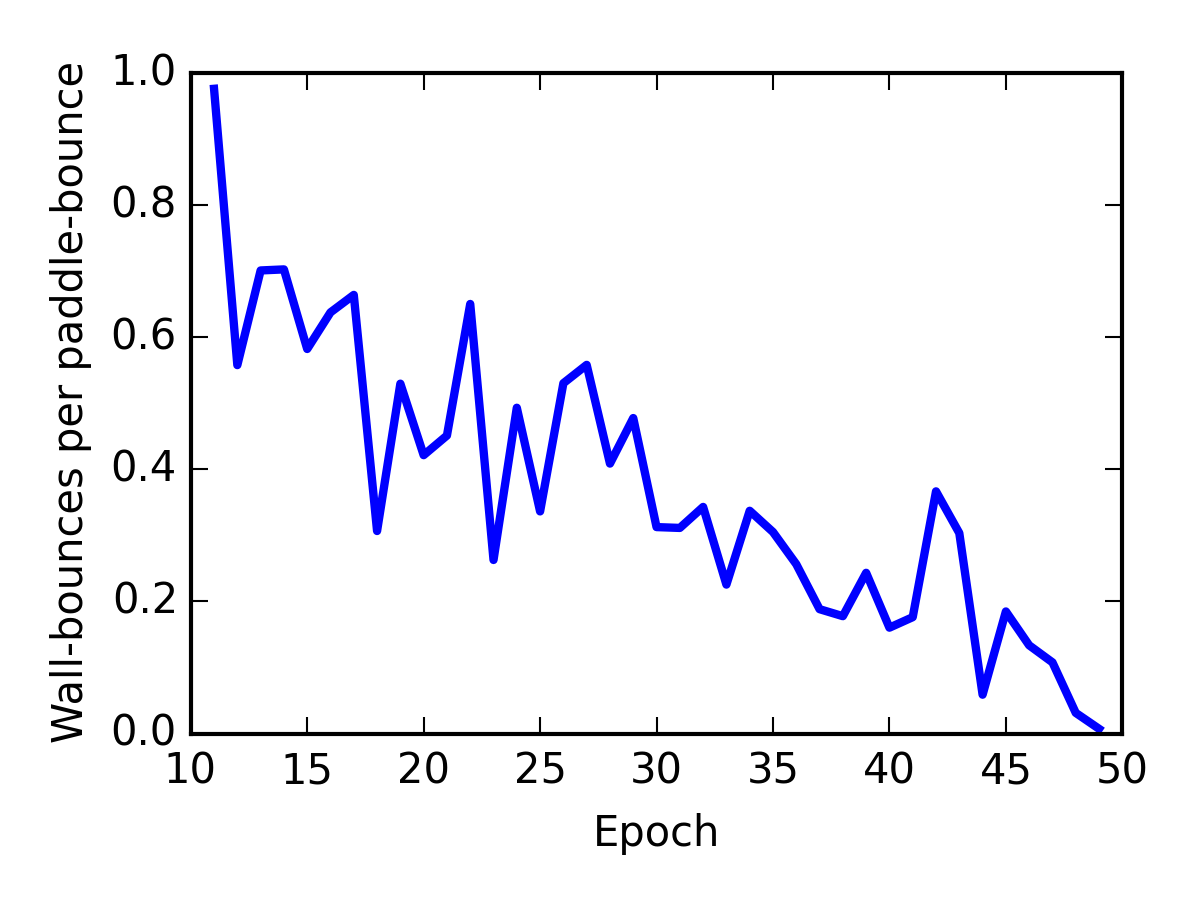}
        \caption{Wall-bounces per paddle-bounce}
        \label{fig:wallbounce_coop}
    \end{subfigure}
    ~ 
    \begin{subfigure}[b]{0.32\textwidth}
        \captionsetup{justification=centering}
        \includegraphics[width=\textwidth]{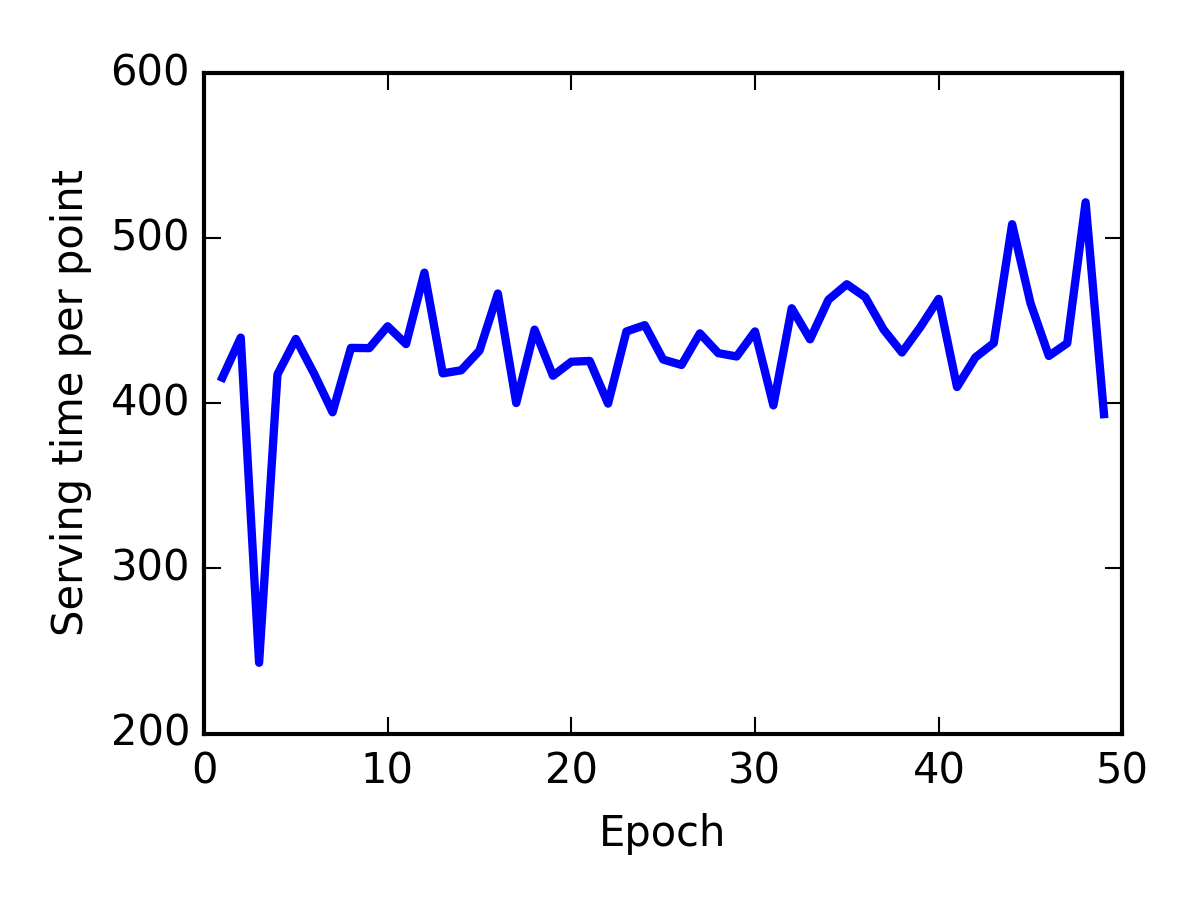}
        \caption{Serving time per point \phantom{-------------------}}
        \label{fig:serving_time_coop}
    \end{subfigure}

    \caption{ Evolution of the behaviour of the collaborative agents during training.  (\subref{fig:sidebounces_coop}) The number of paddle-bounces increases as the players get better at reaching the ball. (\subref{fig:wallbounce_coop}) The frequency of the ball hitting the upper and lower walls decreases significantly with training. The first 10 epochs are omitted from the plot as very few paddle-bounces were made by the agents and the metric was very noisy. (\subref{fig:serving_time_coop}) Serving time increases - the agents learn to postpone putting the ball into play. Serving time is measured in frames.}
    \label{fig:evolution_COOP}
\end{figure} 
The evolution of the quantitative descriptors of behaviour during cooperative training is shown on \Cref{fig:evolution_COOP}. The emergent strategy after 50 epochs of training can be characterized by three observations: (i) the agents have learned to keep the ball for a long time (\Cref{fig:sidebounces_coop}); (ii) the agents take a long time to serve the ball (\Cref{fig:serving_time_coop}) because playing can only result in negative rewards; and (iii) the agents prefer to pass the ball horizontally across the field without touching the walls (\Cref{fig:wallbounce_coop}). Notice that the frequency of wall-bounces decreases gradually with training, whereas the number of paddle-bounces increases abruptly.

\begin{figure}[h!]
\centering
\includegraphics[width=\textwidth]{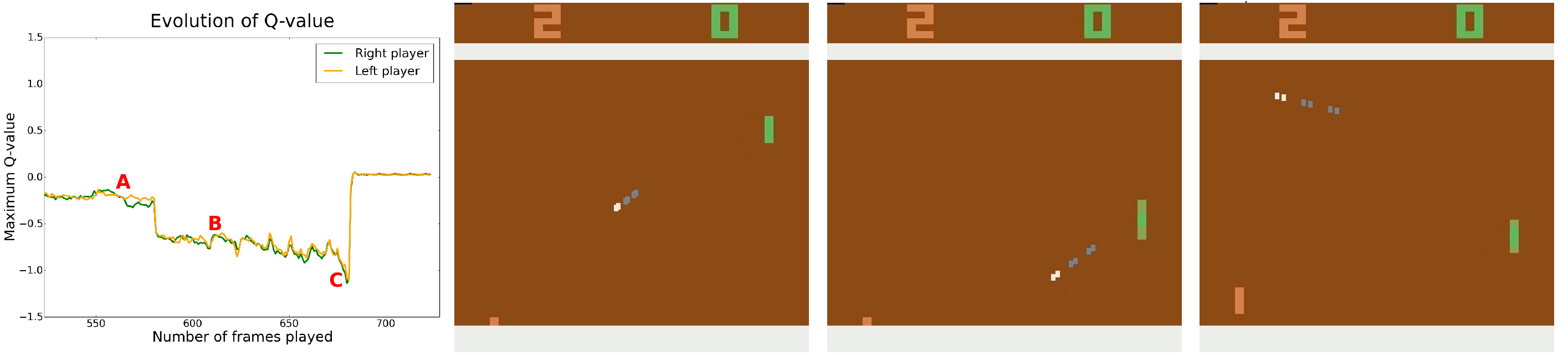}
\caption{Cooperative game. A) The ball is moving slowly and the future reward expectation is not very low - the agents do not expect to miss the slow balls. B) The ball is moving faster and the reward expectation is much more negative - the agents expect to miss the ball in the near future. C) The ball is inevitably going out of play. Both agents' reward expectations drop accordingly. See section \ref{sec:videos} for videos illustrating other game situations and the corresponding agents' Q-values.}
\label{fig:collaborative_game}
\end{figure}

On \Cref{fig:collaborative_game} an exchange between collaborative agents is illustrated. Just like the competitive agents, the collaborative agents learn that the speed of the ball is an important predictor of future rewards - faster balls increase the risk of mistakes. The clear drop in the predicted Q-values in situation B compared to situation A is caused by the ball travelling faster in situation B.

\begin{figure}[H]
\centering
\includegraphics[width=0.55\textwidth]{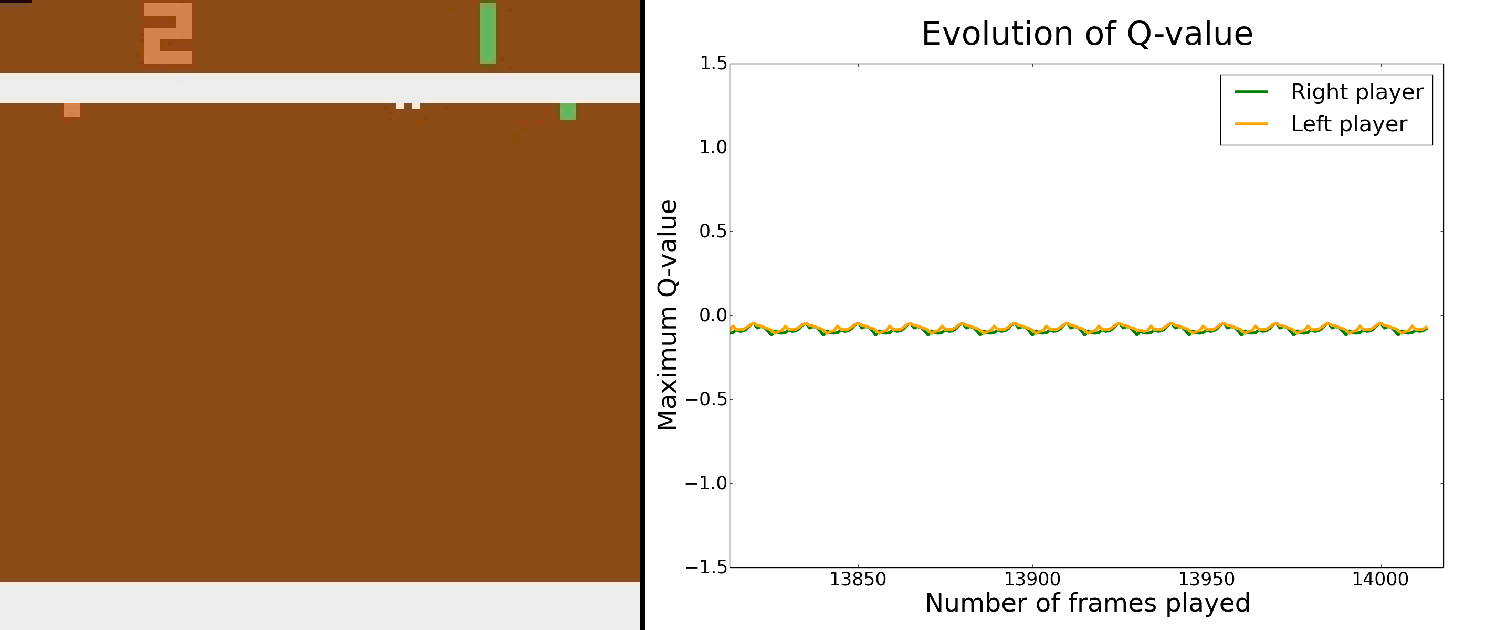}
\caption{In the cooperative setting the agents sometimes reach a strategy that allows them to keep the ball in the game for a very long time.}
\label{fig:cooperative_horizontal}
\end{figure}

In the exchange illustrated on \Cref{fig:collaborative_game} the agents eventually miss the ball. In some exchanges, however, the players apply a coordinated strategy where both agents place themselves at the upper border of the playing field and bounce the ball between themselves horizontally (\Cref{fig:cooperative_horizontal}).
\pagebreak

Whenever a random action takes them away from the edge, they move back towards the edge in the next time step. Note that being at the edge of the field minimizes the effect of random actions - random movements to only one of the two directions are possible. Arriving to the stable situation happens only in some exchanges and might depend on the way the ball is served. It is made harder by the fact that agents choose actions every 4th frame.

See section \ref{sec:videos} for a video illustrating the evolution of the agents' learning progression towards the final converged behaviour and coordinated strategy.

    ~ 

\subsection{Progression from Competition to Collaboration}
Besides the two cases described above, we also ran a series of simulations with intermediate rewarding schemes. Our aim here is to describe the transition from competitive to collaborative behaviour when the immediate reward received for scoring a point ($\rho$) is decreased. 

\begin{figure}[h]
    \centering
    \begin{subfigure}[b]{0.32\textwidth}
        \captionsetup{justification=centering}
        \includegraphics[width=\textwidth]{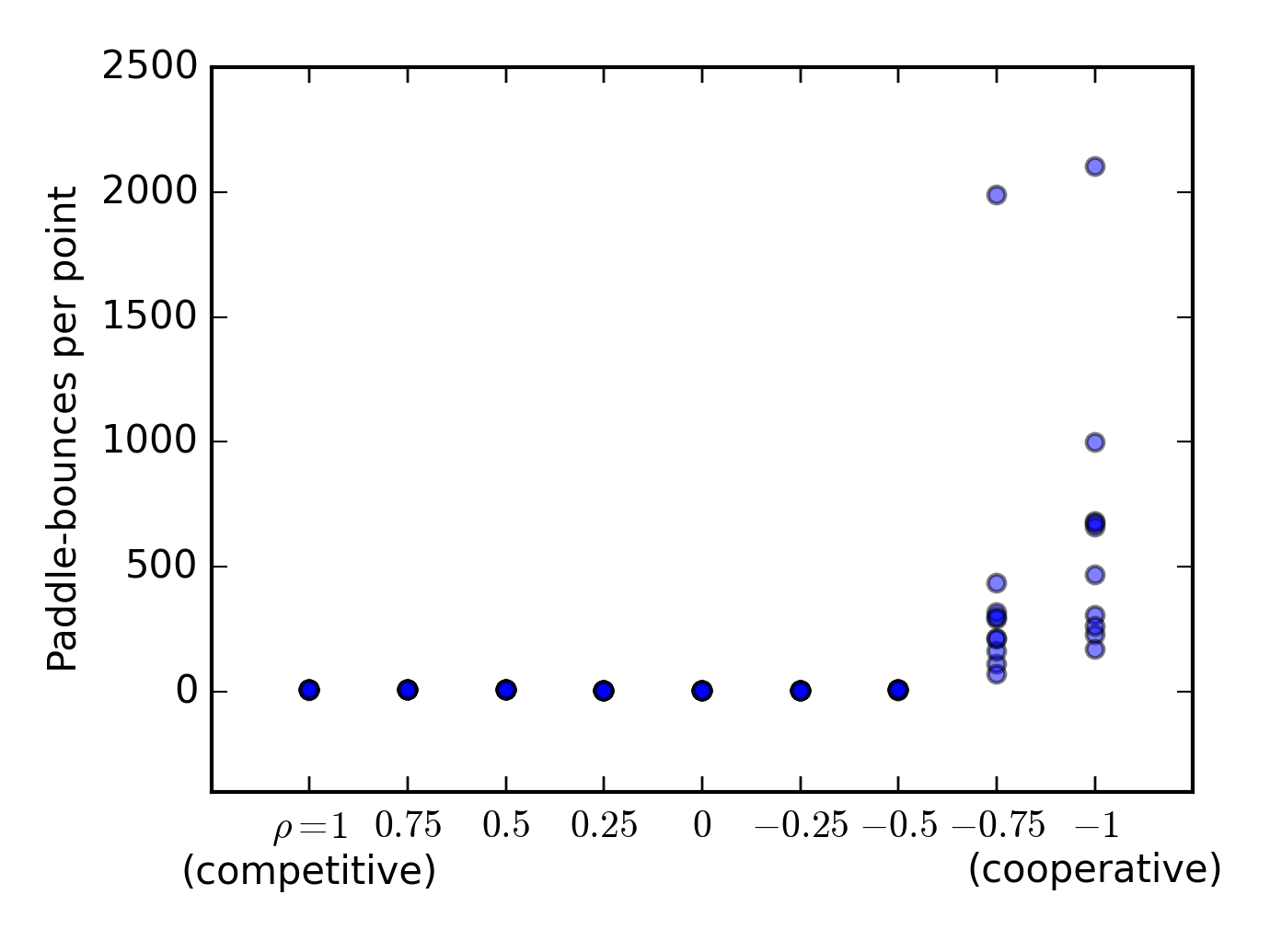}
        \caption{Paddle-bounces per point \phantom{--------------------}}
        \label{fig:sidebounces}
    \end{subfigure}
    ~ 
    \begin{subfigure}[b]{0.32\textwidth}
        \captionsetup{justification=centering}
        \includegraphics[width=\textwidth]{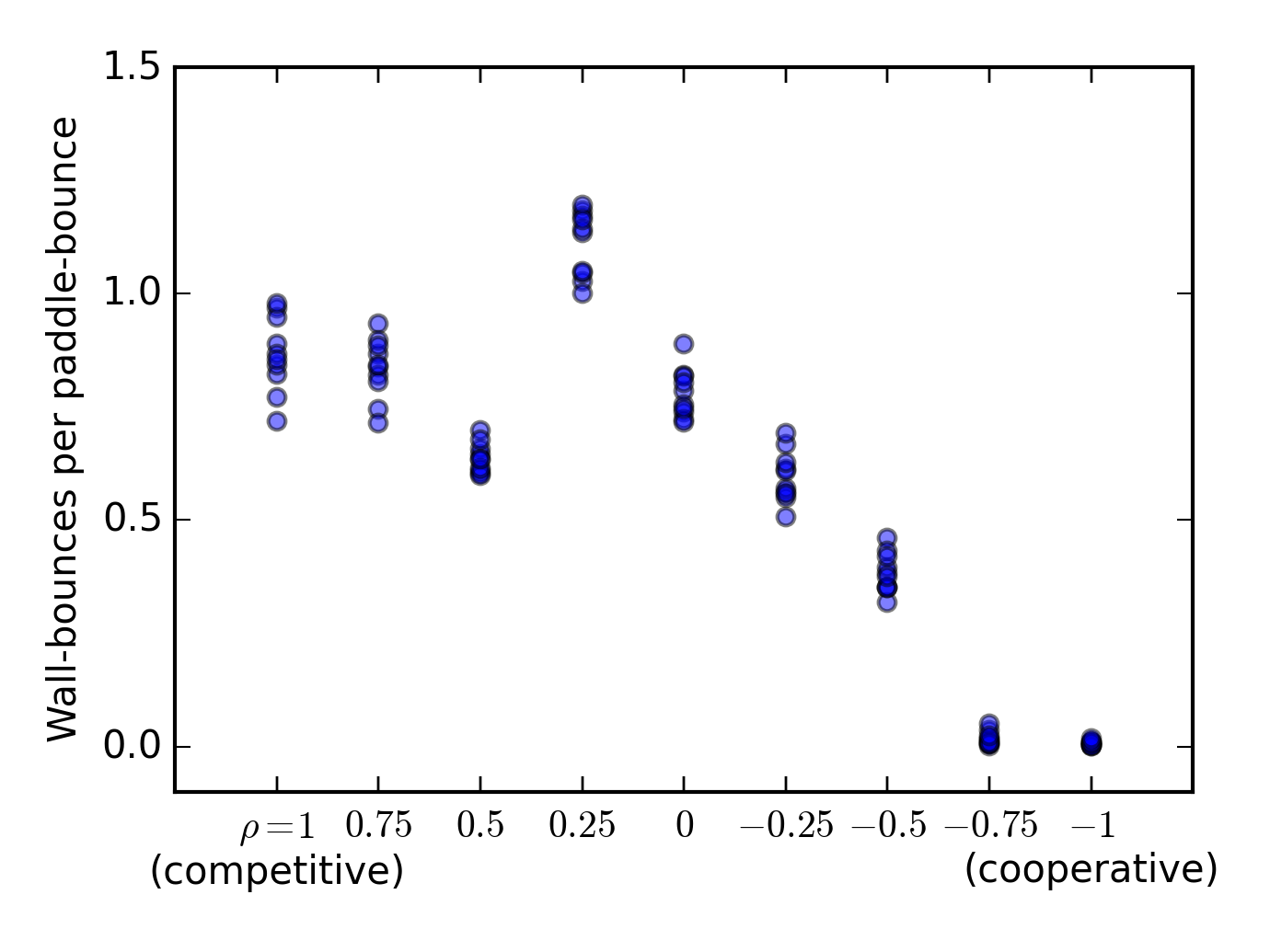}
        \caption{Wall-bounces per paddle-bounce}
        \label{fig:wallbounce}
    \end{subfigure}
    ~ 
    \begin{subfigure}[b]{0.32\textwidth}
        \captionsetup{justification=centering}
        \includegraphics[width=\textwidth]{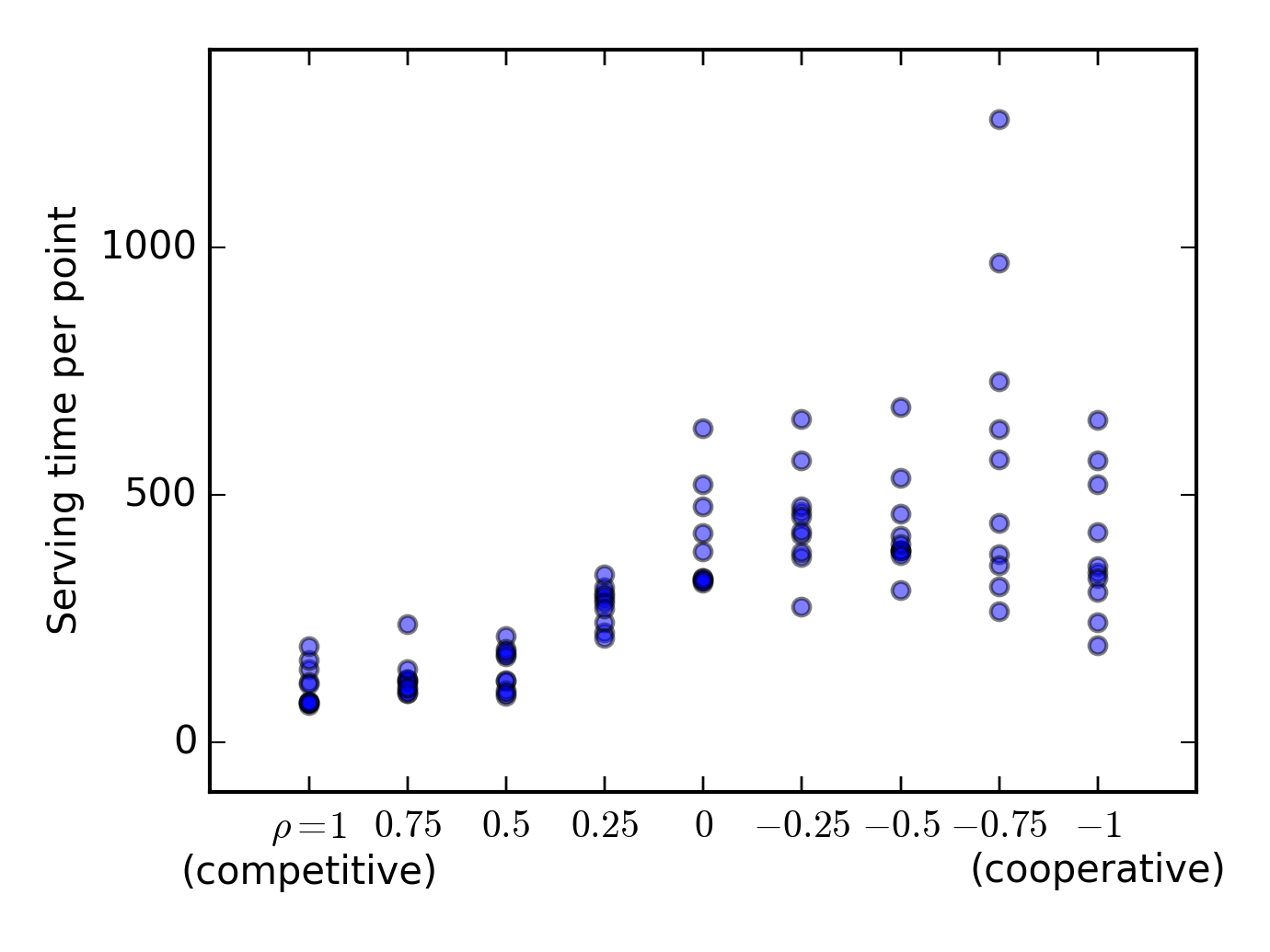}
        \caption{Serving time per point \phantom{--------------------}}
        \label{fig:serving_time}
    \end{subfigure}
    
    \caption{Progression of behavioural statistics when passing from competitive to collaborative rewarding schemes. (\subref{fig:sidebounces}) The number of touches increases when the agents have a strong incentive to collaborate. (\subref{fig:wallbounce}) Forcing the agents to collaborate decreases the amount of angled shots that bounce off the walls before reaching the opposite player. (\subref{fig:serving_time}) Serving time is significantly shorter when agents receive positive rewards for scoring. }
    \label{fig:metrics}
\end{figure}

On \Cref{fig:metrics}, the quantitative behavioural metrics are plotted for decreasing values of $\rho$ in order to give a better overview of the trends. Table \ref{tab:metrics} summarises these results numerically. The statistics are collected after agents have been trained for 50 epochs and are averaged over 10 game runs with different random seeds.




For $\rho=1$, $\rho=0.75$, and $\rho=0.5$ the number of paddle-touches per point stays the same, indicating that the learning algorithm is not very sensitive to the exact size of the reward in that range. However, we observe a significant decrease in the number of touches when we lowered $\rho$ from 0.5 to 0.25 (\Cref{fig:sidebounces}, \Cref{tab:metrics}).

Also the number of wall-bounces per paddle-bounces is the same for reward values 1 and 0.75 (\Cref{fig:wallbounce}, \Cref{tab:metrics}). Lowering $\rho$ to 0.5, the wall-bounces become significantly less frequent indicating some change of behaviour that was not captured by the number of touches. Interestingly, $\rho=0.25$ does not continue the decreasing trend and instead results in the highest number of wall-bounces per paddle-bounce among all the tested rewarding schemes.

Also, for $\rho>0$, weaker positive reward leads to a longer serving time (\Cref{fig:serving_time}, \Cref{tab:metrics}). Let us remind that to restart the game, the agent who just scored has to send a specific command (fire). If the agent never chooses to send this command, the game may still restart via actions randomly selected by the $\epsilon$-greedy exploration. In such case serving takes on average a bit more than 400 frames. This means that the agents trained with $\rho > 0$ actively choose to restart the game at least in some game situations.

Conversely, in all the rewarding schemes with $\rho \leq 0$, the agents learn to avoid relaunching the ball. The average serving times are around 400 frames and correspond to those obtained when agents only start the game due to randomly chosen actions (\Cref{fig:serving_time}, \Cref{tab:metrics}).

For $\rho \leq 0$ we also observe that increasingly negative rewards leads to keeping the ball alive longer (\Cref{fig:sidebounces}, \Cref{tab:metrics}) and to hitting the walls less often (\Cref{fig:wallbounce}). The increased dispersion in statistics for $\rho=-1$ and $\rho=-0.75$ is the result of agents sometimes reaching the strategy of keeping the ball infinitely.
 



\begin{table}[H]
\renewcommand{\arraystretch}{1.5}
\centering
\begin{tabular}{>{\raggedright\arraybackslash}m{33mm} | >{\raggedleft\arraybackslash}m{28mm} >{\raggedleft\arraybackslash}m{25mm} >{\raggedleft\arraybackslash}m{28mm}} 

    \multicolumn{1}{>{\centering\arraybackslash}m{33mm}|}{Agent} &
    \multicolumn{1}{>{\centering\arraybackslash}m{30mm}}{Average paddle-bounces per point} &
    \multicolumn{1}{>{\centering\arraybackslash}m{30mm}}{Average wall-bounces per paddle-bounce} &
    \multicolumn{1}{>{\centering\arraybackslash}m{28mm}}{Average serving time per point} 
    \\\hline
    
    Competitive $\rho=1$ & $ 7.15 \pm 1.01 $ & $ 0.87 \pm 0.08 $ & $ 113.87 \pm 40.30 $ \\
    Transition $\rho=0.75$ & $ 7.58 \pm 0.71 $ & $ 0.83 \pm 0.06 $ & $ 129.03 \pm 38.81 $ \\
    Transition $\rho=0.5$ & $ 6.93 \pm 0.49 $ & $ 0.64 \pm 0.03 $ & $ 147.69 \pm 41.02 $ \\
    Transition $\rho=0.25$ & $ 4.49 \pm 0.43 $ & $ 1.11 \pm 0.07 $ & $ 275.90 \pm 38.69 $ \\
    Transition $\rho=0$ & $ 4.31 \pm 0.25 $ & $ 0.78 \pm 0.05 $ & $ 407.64 \pm 100.79 $ \\
    Transition $\rho=-0.25$ & $ 5.21 \pm 0.36 $ & $ 0.60 \pm 0.05 $ & $ 449.18 \pm 99.53 $ \\
    Transition $\rho=-0.5$ & $ 6.20 \pm 0.20 $ & $ 0.38 \pm 0.04 $ & $ 433.39 \pm 98.77 $ \\
    Transition $\rho=-0.75$ & $ 409.50 \pm 535.24 $ & $ 0.02 \pm 0.01 $ & $ 591.62 \pm 302.15 $ \\
    Cooperative $\rho=-1$ & $ 654.66 \pm 542.67 $ & $ 0.01 \pm 0.00 $ & $ 393.34 \pm 138.63 $ \\

\end{tabular}
\caption{Behavioural statistics of the agents as a function of  their incentive to score.}
\label{tab:metrics}
\end{table}
\vfill
\pagebreak

%% file: discussion.tex

%





Multiagent reinforcement learning has an extensive literature in the emergence of conflict and cooperation between agents sharing an environment \cite{tan1993multi,claus1998dynamics,busoniu2008comprehensive}. However, most of the reinforcement learning studies have been conducted in either simple grid worlds or with agents already equipped with abstract and high-level sensory perception.


In the present work we demonstrated that agents controlled by autonomous Deep Q-Networks are able to learn a two player video game such as \emph{Pong} from raw sensory data. This result indicates that Deep Q-Networks can become a practical tool for the decentralized learning of multiagent systems living a complex environments.

In particular, we described how the agents learned and developed different strategies under different rewarding schemes, including full cooperative and full competitive tasks. The emergent behavior of the agents during such schemes was robust and consistent with their tasks. For example, under a cooperative rewarding scheme the two \emph{Pong} agents (paddles) discovered the coordinated strategy of hitting the ball parallel to the $x$-axis, which allowed them to keep the ball bouncing between them for a large amount of time. It is also interesting to note that serving time, i.e. the time taken by the agent to launch the first ball in a game, was one of the behavioral variables modified during the learning.

\subsection{Limitations}
We observe that in the fully competitive (zero-sum) rewarding scheme, the agents  overestimate their future discounted rewards. The realistic reward expectation of two equally skillful agents should be around zero, but in most game situations both of our Deep Q-Networks predict rewards near 0.5 (\Cref{fig:competitive_game}, supplementary videos). Overestimation of Q-values is a known bias of the Q-learning algorithm and could potentially be remedied by using the novel Double Q-learning algorithm proposed by \cite{hasselt2010double}. 

In addition, with a more accurate estimation of Q-values we might obtain clear collaborative behaviour already with higher values of $\rho$ (less incentive to collaborate) and already earlier in the training process. 

In this work we have used the simplest adaptation of deep Q-learning to the multiagent case, i.e., we let an autonomous Deep Q-Network to control each agent. In general, we expect that adapting a range of multiagent reinforcement algorithms to make use of Deep Q-Networks will improve our results and pave the way to new applications of distributed learning in highly complex environments.

A larger variety of metrics might have helped us to better describe the behaviour of different agents, but we were limited to the statistics extractable from the game memory. More descriptive statistics such as average speed of ball and how often the ball is hit with the side of the paddle would have required analyzing the screen images frame by frame. While probably useful descriptors of behaviour, we decided that they were not necessary for this preliminary characterization of multiagent deep reinforcement learning.

\subsection{Future Work} 
In the present work we have considered two agents interacting in an environment such as \emph{Pong} with different rewarding schemes leading them towards competition or collaboration. Using other games such as \emph{Warlords} we plan to study how a larger number of agents (up to four) organize themselves to compete or cooperate and form alliances to maximize their rewards while using only raw sensory information. It would certainly be interesting to analyse systems with tens or hundreds of agents in such complex environments. This is currently not feasible with the system and algorithms used here. 




A potential future application of the present approach is the study of how communication codes and consensus can emerge between interacting agents in complex environments without any a priori agreements, rules, or even high-level concepts of themselves and their environment.



%% file: appendices.tex

%
%
\subsection{Evolution of Deep Q-Networks}

\begin{figure}[h] 
    \centering
    \begin{subfigure}[b]{0.45\textwidth}
        \includegraphics[width=\textwidth]{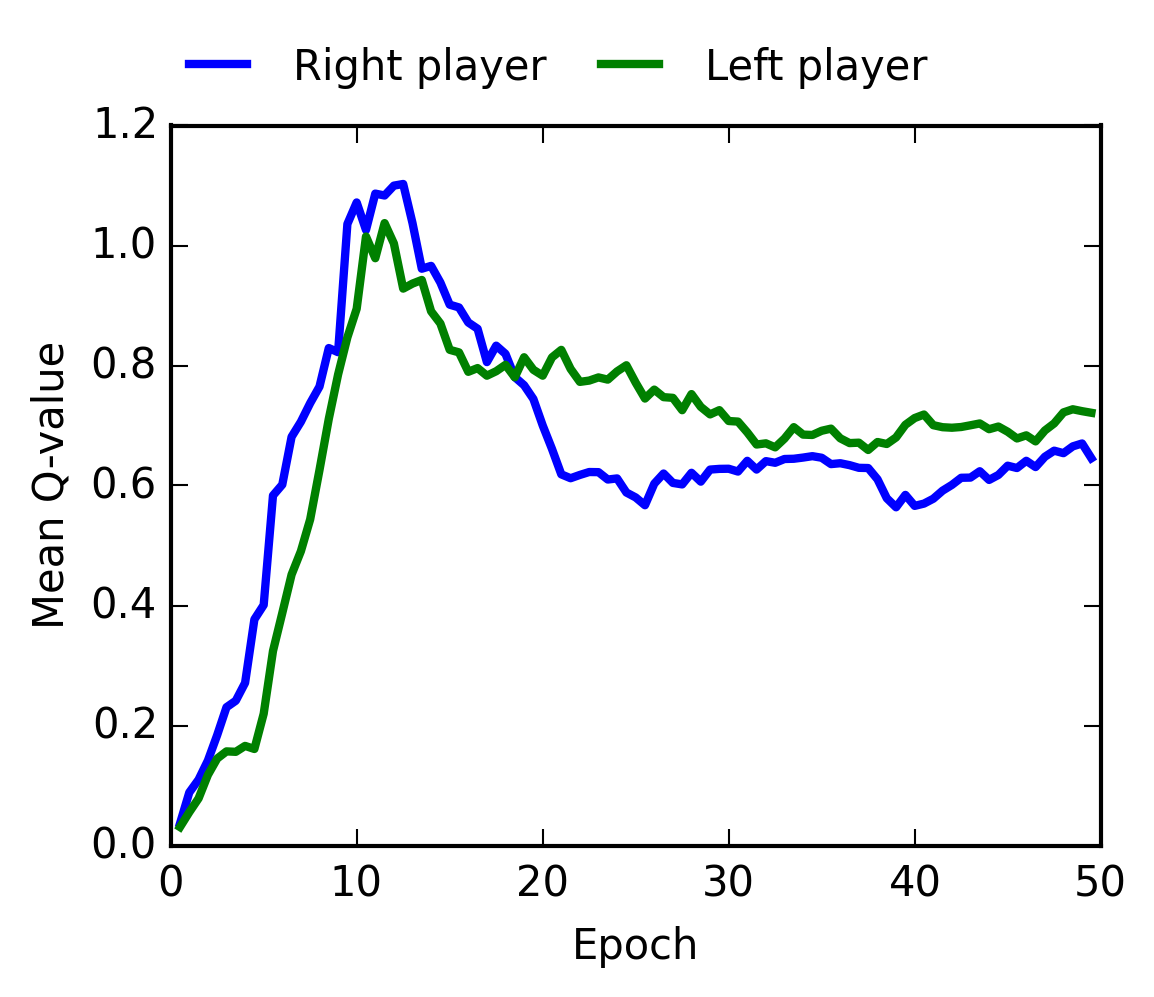}
        \caption{Q-value estimated by the competitive agents}
        \label{fig:meanq_comp}
    \end{subfigure}
    \begin{subfigure}[b]{0.45\textwidth}
        \includegraphics[width=\textwidth]{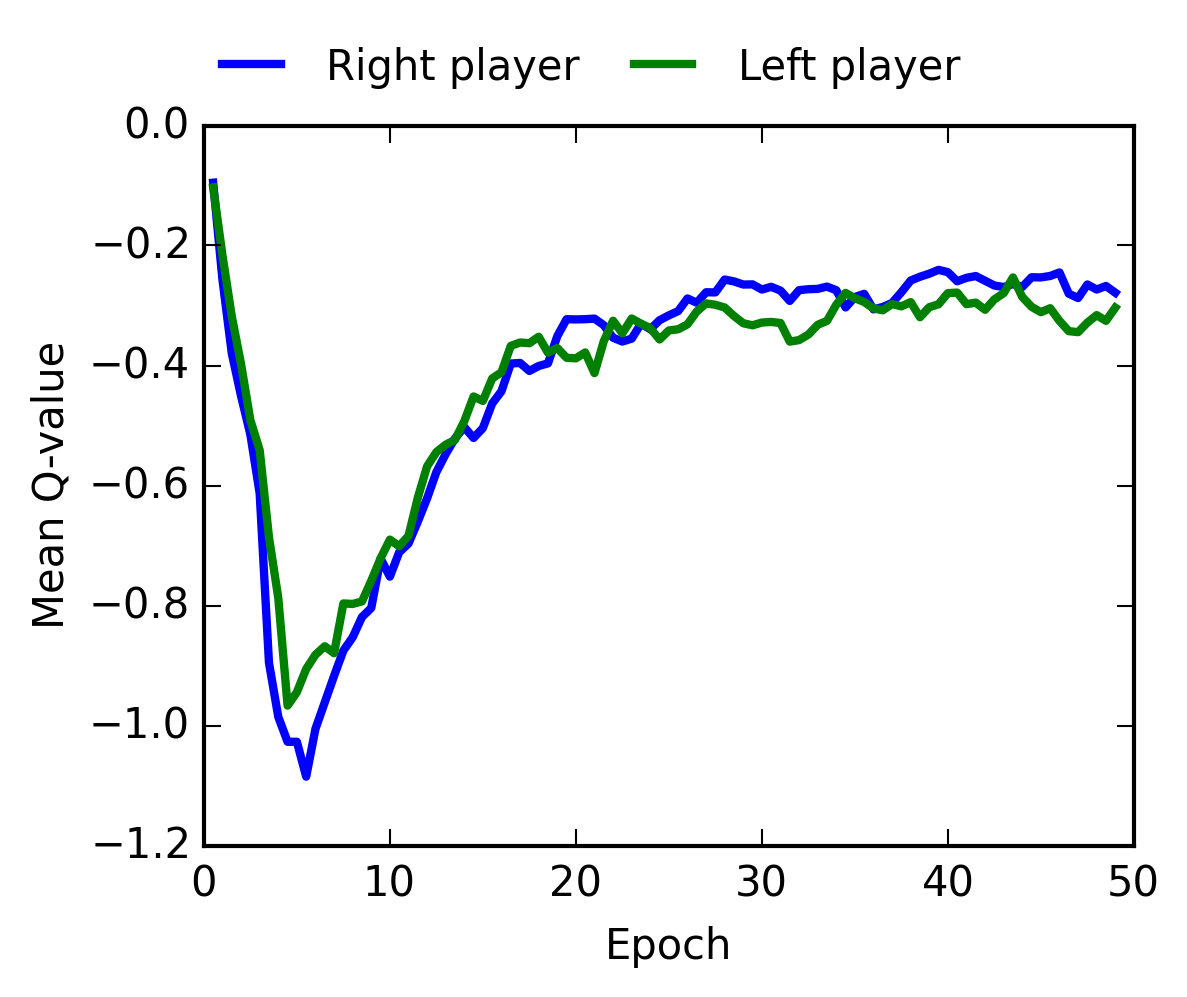}
        \caption{Q-value estimated by the collaborative agents}
        \label{fig:meanq_coop}
    \end{subfigure}
    \caption{Evolution of the Q-value and reward of cooperative and competitive agents over the training time.}
    \label{fig:meanq_reward_all}
\end{figure}

\subsection{Code}
\label{sec:code}
The version of the code adapted to the multiplayer paradigm together with the tools for the visualization can be accessed at our Github repository: \url{https://github.com/NeuroCSUT/DeepMind-Atari-Deep-Q-Learner-2Player}.

%
%
\subsection{Videos}
\label{sec:videos}
Please see the videos illustrating the core concepts and the results of the current work on the dedicated YouTube playlist \url{https://www.youtube.com/playlist?list=PLfLv_F3r0TwyaZPe50OOUx8tRf0HwdR_u}.